\pdfoutput=1

\documentclass[11pt]{article}

\usepackage{emnlp2021}

\usepackage{times}
\usepackage{latexsym}
\usepackage{amsmath}
\usepackage{amssymb}
\usepackage{graphicx}
\usepackage{latexsym}
\usepackage{float}
\usepackage{stfloats}
\usepackage{CJK}
\usepackage{booktabs}

\usepackage[T1]{fontenc}

\usepackage[utf8]{inputenc}

\usepackage{microtype}

\title{Adaptive Bridge between Training and Inference for Dialogue Generation}

\author{
Haoran Xu\textsuperscript{\rm1, 2}\thanks{\; Work done at Data Science Lab, JD.com.},  Hainan Zhang\textsuperscript{\rm3}\thanks{\; Corresponding author.},  \bf{Yanyan Zou\textsuperscript{\rm3}}, \bf{Hongshen Chen\textsuperscript{\rm3}},
\\\bf{Zhuoye Ding\textsuperscript{\rm3}},  \bf{Yanyan Lan\textsuperscript{\rm4}}\\
  \textsuperscript{\rm1}Institute of Computing Technology, Chinese Academy of Sciences, Beijing, China    \\
\textsuperscript{\rm 2} University of Chinese Academy of Sciences, Beijing, China\\
  \textsuperscript{\rm3}Data Science Lab, JD.com, Beijing, China\\
  \textsuperscript{\rm4}Institute of AI Industry Research, Tsinghua University, Beijing, China\\
  \texttt{xuhaoran18s@ict.ac.cn,zhanghainan1990@163.com,zouyanyan6@jd.com}\\
  \texttt{ac@chenhongshen.com,dingzhuoye@jd.com,lanyanyan@tsinghua.edu.cn}\\}

\begin{document}
\maketitle

\begin{abstract}

Although exposure bias has been widely studied in some NLP tasks, it faces its unique challenges in dialogue response generation, the representative one-to-various generation scenario.
In real human dialogue, there are many appropriate responses for the same context, not only with different expressions, but also with different topics.
Therefore, due to the much bigger gap between various ground-truth responses and the generated synthetic response, exposure bias is more challenging in dialogue generation task.
What's more, as MLE encourages the model to only learn the common words among different ground-truth responses, but ignores the interesting and specific parts,  exposure bias may further lead to the common response generation problem, such as ``I don't know'' and ``HaHa?''
In this paper, we propose a novel adaptive switching mechanism, which learns to automatically transit between ground-truth learning and generated learning regarding the word-level matching score, such as the cosine similarity.
Experimental results on both Chinese STC dataset and English Reddit dataset, show that our adaptive method achieves a significant improvement in terms of metric-based evaluation and human evaluation, as compared with the state-of-the-art exposure bias approaches. Further analysis on NMT task also shows that our model can achieve a significant improvement.


\end{abstract}
\section{Introduction}\label{ch:intro}
Auto-regressive models(ARM) are widely used for natural language generation(NLG) tasks, such as machine translation ~\cite{sutskever2014sequence,wu2018adversarial}, dialogue response generation ~\cite{li2017adversarial}, image captioning ~\cite{lin2014microsoft,vinyals2015show} and video description ~\cite{donahue2015long}. 
They utilize the encoder-decoder framework to predict the next token conditioned on the previous tokens, and minimize the cross-entropy between the generation and ground-truths as their objective function. Specifically, at training time, the ground-truth is utilized as the previous tokens, which forces the model directly to learn the distribution of ground truths. But at inference, the previous tokens come from the ARM decoder itself, which is different from the input distribution at training time.

\begin{CJK*}{UTF8}{gbsn}
\begin{table}
\centering
\scriptsize
\newcommand{\tabincell}[2]{\begin{tabular}{@{}#1@{}}#2\end{tabular}}
\begin{tabular}{ll} 
\toprule
\multicolumn{2}{c}{Dialogue 1}\\
\toprule
\hspace{0.5em} context &  \tabincell{l}{听说广州已成避暑胜地}\\ 
&  \tabincell{l}{( I heard that Guangzhou has become a summer resort)}\\ 
response1 &  \tabincell{l}{确实，这边很凉快。( Indeed, it's cool here. )}\\ 
response2 &  \tabincell{l}{晚上睡觉都没开风扇了。}\\ 
&  \tabincell{l}{( There is no need to turn on the fan at night. )}\\ 
\toprule
\multicolumn{2}{c}{Dialogue 2}\\
\toprule
\hspace{0.5em} context &  \tabincell{l}{哈哈，看看可爱的小猫咪( Ha ha, look at this lovely kitten )}\\ 
response1 &  \tabincell{l}{这是什么品种的猫哇？ \textcolor{red}{好可爱，我也想要}}\\ 
&  \tabincell{l}{( What kind of cat is this? \textcolor{red}{So cute, I want it too.} )}\\ 
response2 &  \tabincell{l}{\textcolor{red}{好想要一只这样的猫}，可以陪我儿子玩}\\ 
&  \tabincell{l}{( \textcolor{red}{I really want a cat like this} to play with my son. )}\\ 
response3  &  \tabincell{l}{哇，\textcolor{red}{好可爱哇，}我屋也有一只这样的小猫 }\\ 
    &  \tabincell{l}{( Wow, \textcolor{red}{It's so cute,} I have a kitten like this in my house. )}\\
\bottomrule
\end{tabular}
\caption{The two Dialogues in STC dataset, and the red part of responses in Dialogue 2 are the common words.} 
\label{tb:examples}
\end{table}
\end{CJK*}

Although this discrepancy, named exposure bias, has been studied in some classic NLG tasks, such as neural machine translation(NMT) ~\cite{bengio2015scheduled,venkatraman2015improving,zhang-etal-2019-bridging}, it faces its unique challenges in dialogue response generation, the representative one-to-various generation scenario.
In human dialogue, given the context, people can reply many relevant and appropriate responses, not only with various expressions but also with different topics. Take the Dialogue 1 in Table ~\ref{tb:examples} as an example, given the context ``I heard that Guangzhou has become a summer resort'', the response 1 and response 2 are in the same topic but with different tokens. In this various expression situation, like NMT task, data distribution and model distribution are easy to fit, relatively, even with exposure bias problem. However, in different topics situation, data distribution is often different from the model, because it is too divergent and covers various word distribution of each topic. Through our data analysis, we find that in dialogue generation task, the various ground-truth responses and the generated sentences have a bigger gap than in NMT tasks. We calculate the overlap measures at word-level and semantic-level, i.e., BLEU and cosine similarity, between the generated sentence and the ground-truth sentences. The results show that on NMT WMT'14 dataset, the BLEU and similarity are 27.38 and 0.96, respectively, while on dialogue Reddit dataset, the BLEU and similarity are 2.17 and 0.81, respectively. We can see that the overlap measures of the dialogue generation task are significantly lower than that of the NMT task, which indicates the severity of the exposure bias problem in the dialogue generation. 

What's more, as Maximum Likelihood Estimation(MLE) encourages the model to only learn the common words among different ground-truth responses, but ignores the interesting and specific parts, exposure bias may aggravate the common response problem of the generation model, due to the strict matching between the generated response and the ground-truth responses. Take the Dialogue 2 in Table ~\ref{tb:examples} as an example, the response 1 is ``What kind of cat is this? So cute, I want it too.'', the response 2 is ``I really want a cat like this to play with my son. '' and the response 3 is ``Wow, it's so cute, I have a kitten like this in my house.''. If we train the model with word-level strict matching between the generated response and the ground-truth, it can only learn the common words, i.e.,  ``So cute, I want it'', but ignore the specific parts, i.e., ``What kind of cat is this?''. Therefore, it is beneficial to improve the strict matching mechanism for the dialogue generation task.


In this paper, we propose a novel Adaptive switch mechanism as a Bridge(AdapBridge), which introduces the generator distribution to the training phase and learns to automatically transit between the ground-truth learning and the generated learning, with respect to the word-level matching scores, such as the cosine similarity. Specifically, at each training step, we calculate the cosine similarity for each generated word with respect to all its ground-truths. If the matching score is bigger than the threshold, the generated word is fed to the decoder, while if lower, the ground-truth is fed for training. The threshold is increasing as the training epoch grows. With this adaptive sampling scheme, the switch mechanism can consider the generation quality of every word, i.e., relevance between the generated word and the ground-truth, to decide whether utilizing the generated learning or not.

We evaluate the proposed models on two public datasets,
i.e. the Chinese STC and the English Reddit dataset. Experimental results show that our models significantly outperform the state-of-the-art exposure bias models with respect to both metric-based evaluations and human judgments. Further analysis on NMT task also shows that our model can achieve a significant improvement.

The main contributions of this paper include:
\begin{itemize}
\item We study the exposure bias problem in dialogue generation task, one of the one-to-various generation scenarios. And find that the exposure bias may further lead to the common response generation problem.
\item We propose the adaptive switch mechanism with word-level matching scores to determine the training input source, in order to resolve the common response problem.
\item We evaluate AdapBridge on two public dialogue datasets and conduct rigorous experiments to demonstrate the effectiveness of our proposed models. Further analysis on NMT task also shows that our model can achieve a significant improvement.
\end{itemize}

\section{Related Work}\label{ch:relate}

This section briefly introduces recent research progresses related to this work in literature.

To solve the exposure bias problem in auto-regressive or seq2seq models ~\cite{sutskever2014sequence,welleck2019neural,holtzman2019curious}, ~\citet{venkatraman2015improving} tried to use data as Demonstrator(DAD) to augment the training set through the tokens predicted by the model, so as to make the training set to meet the test distribution. The method of Scheduled Sampling(SS) proposed by ~\citet{bengio2015scheduled} attempted to randomly sample the previous generated words to replace the ground-truth words for the model input during training time. ~\citet{zhang-etal-2019-bridging} made a further exploration of this method by sampling the previous words with decay not only from word-level oracle but also from the sentence-level oracle with a semantic metric. 
The main idea of this kind of method is to introduce the model's prediction information to its input at training time, and reduce the discrepancy between training and inference to alleviate the exposure bias problem. In comparison to those methods and related ideas ~\cite{qi2020prophetnet,goodman2020teaforn}, our proposed method adaptively determines whether the input words of model during training are ground truth or predicted by scoring each generated word. 

Alternative based on Reinforcement Learning(RL) ~\cite{williams1992simple} methods have been explored for generation tasks, in particular for NMT.
Mixed Incremental Cross-Entropy Reinforce (MIXER)  ~\cite{ranzato2015sequence} leverage  hybrid loss function which combines both cross-entropy and reinforce to directly optimized the metrics used at test time, such as BLEU or ROUGE. There are many other similar works ~\cite{shen2015minimum,yonghui2016bridging,shao2018greedy}. More recently, text generation via Generative Adversarial Networks(GAN) ~\cite{goodfellow2014generative} called Text GANs has attracted of researchers ~\cite{nie2018relgan,zhou2020self,wu2020textgail,scialom2020coldgans}. They framed the problem under the GAN paradigm, which uses the RL-Based ~\cite{williams1992simple} algorithms to get the gradient estimation, as the text generation is discrete. However, both RL and Text GANs cannot be avoided the high variance of gradient estimation caused by sparse rewards, which consequently makes the training process unstable and limits improvements.

Different from traditional methods, our proposed model can adaptively determine whether the current input word is from ground truth or from generation with the word-level matching scores.

\begin{figure}[!t]
    \centering
    \includegraphics[width=0.99\linewidth]{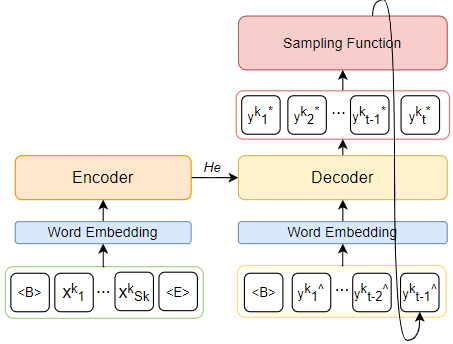}
    \caption{The illustration of our AdapBridge Model.}
    \label{fig:randomsample}
\end{figure}

\begin{figure*}[!t]
    \centering
    \includegraphics[width=1\linewidth]{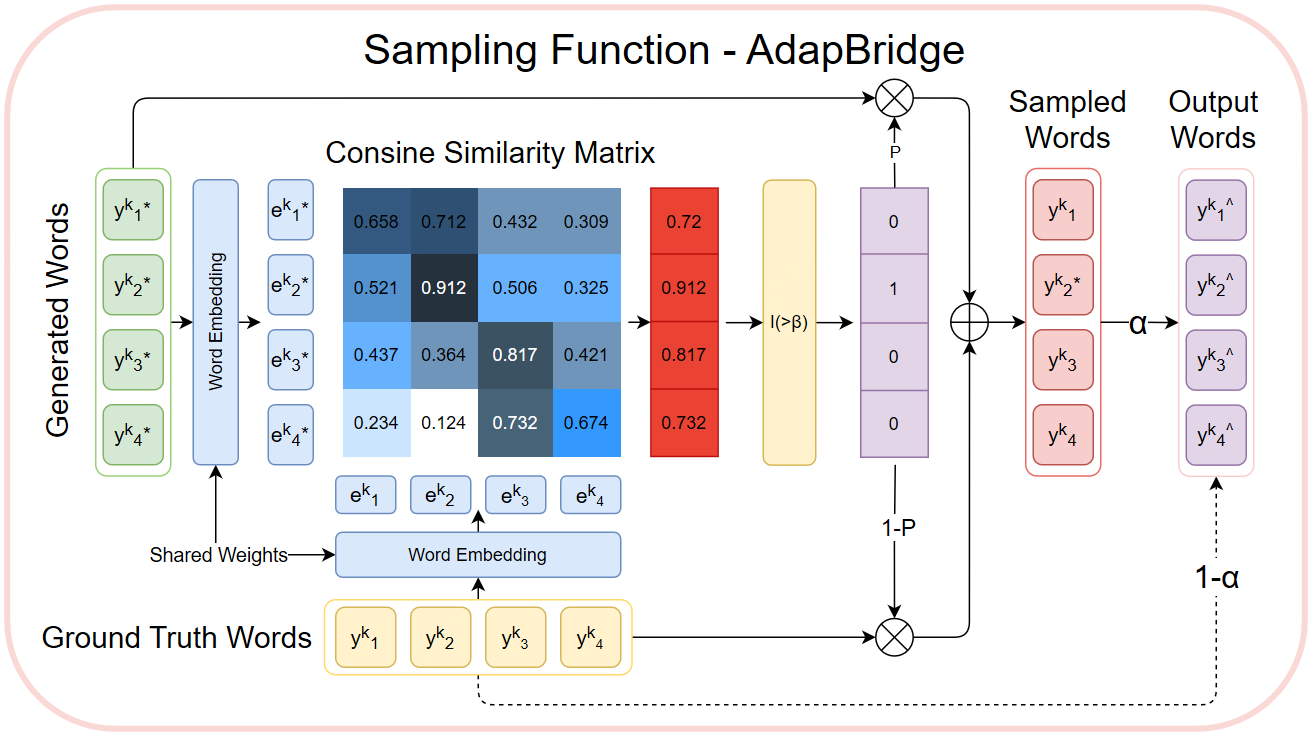}
    \caption{The Illustration of the sampling function Adapt-Bridge. The word embedding shares weights with encoder and decoder of model showed in Figure  ~\ref{fig:randomsample}. $I(>\beta)$is a Indicator function, if input is upper $\beta$, output will be 1, otherwise, output will be 0. $\alpha$ and $\beta$ are both increasing as the training epochs grows.}
    \label{fig:adaptbridge}
\end{figure*}

\section{Proposed Method}
Given a context sentence $X^k=\{x_{1}^{k},x_{2}^{k},\cdots,x_{S_k}^{k}\}$, and a target response sentence $Y^k=\{y_{1}^{k},y_{2}^{k},\cdots,y_{T_k}^{k}\}$, where $S_k$  and $T_k$ are the word length of context and response, respectively. The dialogue generation model based on sequence-to-sequence (Seq2Seq) ~\cite{sutskever2014sequence} framework, directly models the response probability:
\begin{equation}
    P(Y^k|X^k,\theta) = \prod^{T_k}_{t=1}p(y_t^k|y_{<t}^k,X^k,\theta)
    \label{equ:prob}
\end{equation}
where $\theta$ are the parameters of model and $y_{<t}^k$ denote the previous ground-truth words.
Given a set of training examples $D=\{X^k,Y^k\}_{k=1}^{N}$, the stander training objective is to minimize the negative log-likelihood of all the training data:
\begin{equation}
    \theta=\mathop{argmin}\limits_{\theta}\{ L (\theta)\}
\end{equation}
\noindent where
\begin{equation}
    \begin{split}
    L(\theta)&=\sum_{k=1}^{N}-logP(Y^k|X^k,\theta)\\
    &=\sum_{k=1}^{N}\sum_{t=1}^{T_k}{-log}p(y_t^k|y_{<t}^k,X^k,\theta)
    \end{split}
\end{equation}
where N is the number of training examples.


Different from training time, during inference, the probability of each target word $p(y_t^k|y_{<t}^k,X^k,\theta)$ in Equation ~\ref{equ:prob} is conditioned on the previous generated words ${y_{<t}^{k}}^*$ rather than the ground-truth  $y_{<t}^{k}$ , as the ground truth words are not available in real inference time. This discrepancy is called exposure bias.


\subsection{AdapBridge}
The architecture of our model is illustrated in Figure ~\ref{fig:randomsample}. 
We first define a sampling function:
\begin{equation}
    {y_{<t}^{k}}^{\wedge}=\mathbf{SamFun}({y_{<t}^{k}},{y_{<t}^{k}}^{*})
    \label{equ:samplfunciton}
\end{equation}
where ${y_{<t}^{k}}$ and ${y_{<t}^{k}}^{*}$ are the inputs of sampling function, representing the ground truth words and generated words, respectively, and ${y_{<t}^{k}}^{\wedge}=\{{y_{1}^{k}}^{\wedge},{y_{2}^{k}}^{\wedge}\cdots,{y_{t-1}^{k}}^{\wedge}\}$ denotes the inputs of decoder after sampled by the sampling function at $t-th$ time step, which may contain both ground truth and generated words.

In this framework, to predict $t-th$ target word ${y_{t}^{k}}^{*}$, we can follow those steps:
\begin{itemize}
    \item Decoders predict $t-1$ words ${y_{<t}^{k}}^{*}$ as the previous generated words.
    \item Use the sequences ${y_{<t}^{k}}^{*}$ and ${y_{<t}^{k}}$(ground truth words) as inputs of \textbf{{SamFun}}(Equation ~\ref{equ:samplfunciton}), and get the outputs ${y_{<t}^{k}}^{\wedge}$ of this function.
    \item Replace the inputs ${y_{<t}^{k}}$ of model in Equation ~\ref{equ:prob} with ${y_{<t}^{k}}^{\wedge}$, then predict the t-th word ${y_{t}^{k}}^{*}$.
\end{itemize}
The \textbf{SamFun} can be any function, i.e. random sampling. The process is to replace the corresponding ground-truth words in ${y_{<t}^{k}}$ with the generated words ${y_{<t}^{k}}^{*}$. We propose a novel \textbf{SamFun} called AdapBridge, which can be seen in Figure \ref{fig:adaptbridge}.

\begin{table}[!t]
\centering
\begin{tabular}{rl}
\toprule
\multicolumn{2}{l}{\textbf{Algorithm 1} AdapBridge }\\
\toprule
\multicolumn{2}{l}{\textbf{Input:}}\\
\multicolumn{2}{l}{\hspace{2em}Sequence of generated words ${y_{<t}^{k}}^{*}$;}\\
\multicolumn{2}{l}{\hspace{2em}Sequence of ground-truth words $y_{<t}^{k}$;}\\
\multicolumn{2}{l}{\hspace{2em}Word embedding of model with size of }\\
\multicolumn{2}{l}{\hspace{2em}shared vocabulary; Number of epoch $n$.}\\
\multicolumn{2}{l}{\textbf{Output:}}\\
\multicolumn{2}{l}{ \hspace{2em}Inputs of decoder after sampled ${y_{<t}^{k}}^{\wedge}$}\\
1:& Initialize $P\leftarrow \{p_1,p_2,\cdots,p_{|{e_{<t}^{k}}^{*}|}\}$ \\
&Calculate $\alpha$ and $\beta$  with $n$\\
&Get a random number $m$ between [0,1].\\
2:& \textbf{if} $m<\alpha$ \textbf{then} \\
3:&\hspace{2em} ${e_{<t}^{k}}^{*}\leftarrow \textbf{E}({y_{<t}^{k}}^{*})$, $e_{<t}^{k}\leftarrow \textbf{E}(y_{<t}^{k})$ \\
4:&\hspace{2em} \textbf{for} $i=1, \cdots, |{{e_{<t}^{k}}^{*}}|$  \textbf{do} \\
5:&\hspace{3em} \textbf{for} $j=1, \cdots, |{e_{<t}^{k}}|$  \textbf{do} \\
6:&\hspace{4em} $\mathbf{SimMat}(i,j)\leftarrow\frac{ {{e_{i}^{k}}^{*}} \cdot  e_{j}^{k} }{|{e_{i}^{k}}^{*}||e_{j}^{k}|}$\hspace{1em}\\
7:&\hspace{3em} \textbf{end for}\\
8:&\hspace{3em} $s_{i}\leftarrow\mathop{\mathbf{MaxSim}}\limits_{j\in 1 ,\cdots, |{e_{<t}^{k}}|}(\mathbf{SimMat}(i,j))$ \\
9:&\hspace{3em} $p_i \leftarrow I_{>\beta}(s_{i})$\\
10:&\hspace{2em} \textbf{end for}\\
11:&\hspace{2em}${y_{<t}^{k}}^{\wedge}\leftarrow {y_{<t}^{k}}^{*}\otimes P+y_{<t}^{k}\otimes(1-P)$\\
12:& \textbf{else}\\
13:&\hspace{2em} ${y_{<t}^{k}}^{\wedge}\leftarrow y_{<t}^{k}$\\
14:& \textbf{end if}\\
15:& \textbf{return} ${y_{<t}^{k}}^{\wedge}$\\
\toprule
\end{tabular}
\caption{Algorithm of AdapBridge}
\label{tab:Algorithm}
\end{table}

The main idea of AdapBridge is simple: we first use the model to generate all words with Equation ~\ref{equ:prob}, and then compute the pairwise cosine similarity between the generated words and the ground truth. If the generated word is learned good enough (similar to the ground truth or a synonym), the max cosine similarity of this word will be close to 1 and upper a threshold $\beta$ showed in Figure ~\ref{fig:adaptbridge}. Therefore, we can use this generated word to replace the ground truth word, which introduces the generator distribution to the training phase. The summary of the algorithm is illustrated in Table ~\ref{tab:Algorithm}. 
\subsection{Sampling with Increase }
The threshold $\alpha$ and $\beta$ determine the frequency of sampling function and the similarity between generated words with ground truth, respectively. Note that, when $\alpha=0$, the training type is same as before, while when $\alpha=1$, the model is trained as inference. If $\alpha$ is set too low($\approx 0$), the inputs of decoder will almost be ground-truth, and will not be able to cope with the unknown words predicted in the reference. On the other hand, if $\alpha$ is set too high($\approx 1$) at the beginning of training, the model will yield tokens randomly, because the model is not well trained, which may lead to slow convergence. Similarity, because model can not generate high cosine similarity score at the beginning, it is necessary to set the $\beta$ low to ensure that part of ground truth words can be replaced by generated words, and increase its value as the training steps or epochs growing. In this sense, $\alpha$ and $\beta$ should be both increase with the training time grows. Note that, the probability $\alpha$ in our method determine whether to execute the transition mechanism. And the the threshold $\beta$ determine which words in the generated sentence should be replaced, with respect to the cosine similarity score. 

\subsubsection{Increase Function of $\alpha$}
We define $\alpha$ with an increase function dependent on the number of training epochs $n$:
\begin{equation}
    \alpha = 1 - \frac{k}{k+exp((n-w)/k)}
    \label{equ:alpha}
\end{equation}
where $k\ge1$ is hyper-parameter, which determines the speed of convergence. In addition, we add a parameter $w$, which makes $\alpha$  close to 0 during the first $w$ epochs of training. It is usually set to half number of all training epochs to ensure that the model is trained enough to generate reasonable words. The curve of this function can be seen in Figure ~\ref{fig:curveof}.

\subsubsection{Increase Function of $\beta$}

\begin{figure}[t]
    \centering
    \includegraphics[width=0.9\linewidth]{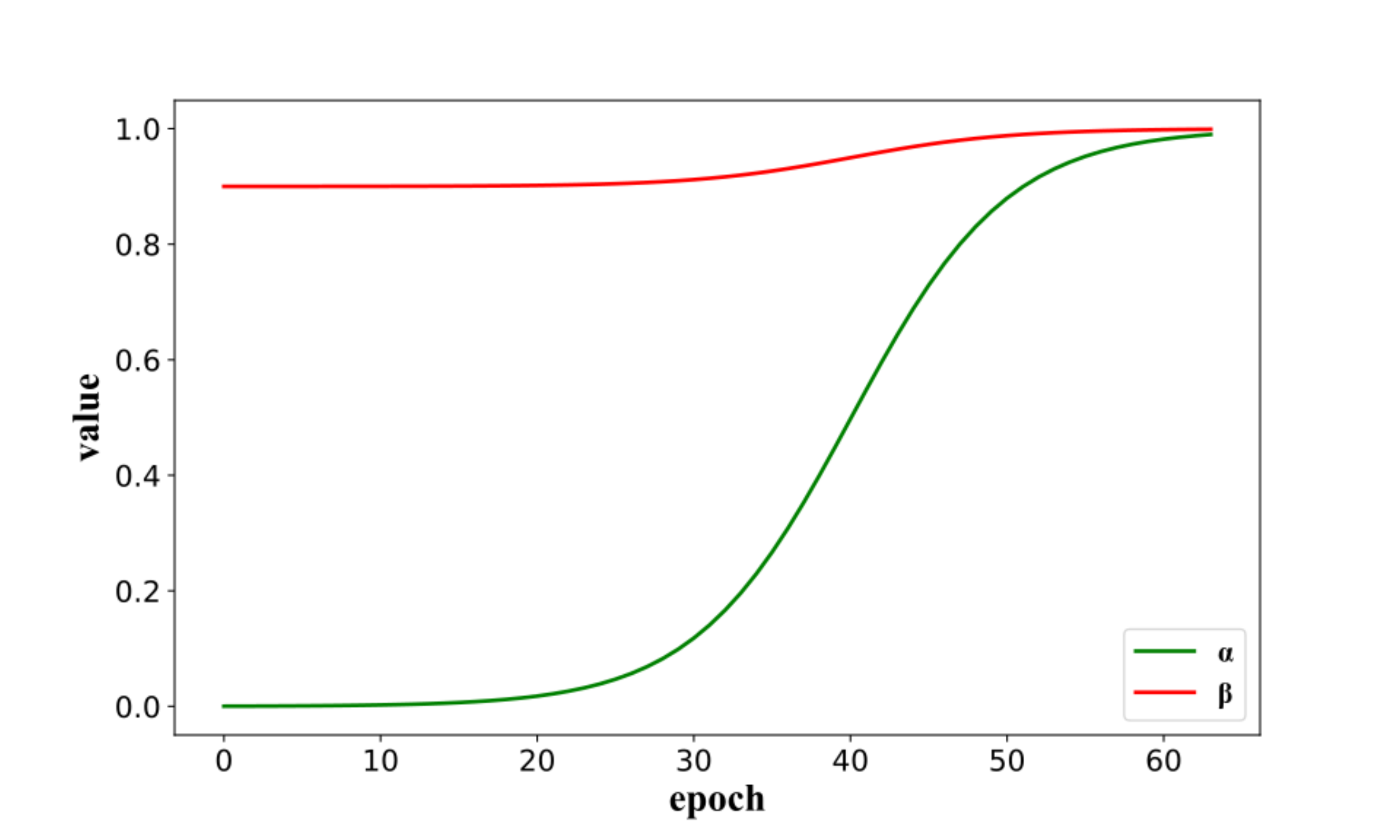}
    \caption{The curve of $\alpha$ and $\beta$. Where $k=5,w=32$ in Equation ~\ref{equ:alpha}, and $\gamma=0.9$ in Equation ~\ref{equ:beta}}
    \label{fig:curveofalphabeta}
\end{figure}

At the beginning of training, the words cosine similarity scores generated by the model are generally low, so  reducing $\beta$ appropriately could help. On the other hand, at the end of training, $\beta$ should increase, as the model was already trained good enough to generate meaningful words after the first $w$ epochs. However, $\beta$ can not start from zero as $\alpha$ showed in Figure ~\ref{fig:curveofalphabeta}. Intuitively, even at the beginning, it should have a certain threshold to ensure that the generated words which will replace the ground truth words have a high quality.

We thus propose to use a schedule to increase $\beta$ as a function of $\alpha$ calculated with Equation ~\ref{equ:alpha} , the formula as follows:
\begin{equation}
    \beta = \gamma + (1-\gamma)*\alpha
    \label{equ:beta}
\end{equation}
where $\gamma$ is the lowest similarity score threshold. In the entire training process, one ground truth word can be replaced only when the score of the generated word is at least greater than $\gamma$. The function is strictly monotone increasing the same as the Equation ~\ref{equ:alpha}, and its curve can be seen in Figure ~\ref{fig:curveofalphabeta}.
\section{Experiments}
In this section, we evaluate our proposed model on both Chinese STC and the English Reddit dataset.

\subsection{Experimental Settings}
We first introduce datasets, baselines, parameters setting and evaluation measures.
\subsubsection{Datasets}
We use two public one-to-many single turn dialogue datasets in our experiments. The Chinese Weibo Dataset, named STC\footnote{https://cloud.tsinghua.edu.cn/f/372be4a9994b4124810e/}, consists  4,391,266 , 23,532 and 21,161 dialogue context-response pairs for training, validation and testing sets, respectively. We remove those pairs whose context contains response or contrary, and obtain 4,295,557 , 23,039 and 20,749 pairs for three data sets. The average number of responses corresponding to each context in STC is 19.7. The English Reddit dialogue corpus is extracted from the Reddit comments-posts by the script\footnote{https://github.com/bsantraigi/RedditDialogue}, named Reddit. The original data consists of 6 million dialogues from all even months of 2011. We use the official script to tokenize, and the duplicates and sentences with length less than 3 or longer than 64 are removed. If the number of responses of one context is more than 20, we will randomly select 20 responses and ignore others, and the average number of responses corresponding to each context in Reddit is 6.2. Finally, we randomly split the data to training, validation and testing sets, which contains 1,107,860, 23,183, 12,429 pairs, respectively.
\subsubsection{Baselines and Parameters Setting}\label{ch:setting}
Three baselines are used for comparison, including Transformer-based model~\cite{vaswani2017attention}, Random Sampling with word(RS-word) and sentence(RS-Sentence)~\cite{zhang-etal-2019-bridging}.

For STC, we utilize the Chinese word as input, and set the vocabulary size as 10,599. For Reddit,  context-response pairs are encoded using byte-pair encoding(BPE) ~\cite{sennrichetal2016neural} with vocabularies of 11,527 tokens. For a fair comparison among all baseline models and our model,  the dimension of all word embedding is 512, and beam size in testing is 5. The transformer model has 6 layers in both encoder and decoder, while 8 heads in multi-head attention. All parameters are initialized by the uniform distribution over$ [-0.1, 0.1] $.
We adopt the optimizer Adam ~\cite{kingma2014adam} with $\beta{1}=0.9, \beta_{2}=0.98$ and with a weight decay of $\epsilon = 10^{-8}$ . We set the learning rate as 0.0007 and the maximum tokens of a batch as 8192 with the update frequency 2. We run all models on 4 Tesla P40 GPU cards with PyTorch\footnote{https://github.com/pytorch/fairseq}. The code will be released when this paper is accepted.

\subsubsection{Evaluation Measures}
The evaluation of quantitative metrics and human judgements are used in our experiments. Specifically, quantitative metrics contains traditional metrics, such as PPL and BLEU score ~\cite{papineni2002bleu}, and Distinct ~\cite{li2015diversity} metric which is recently proposed to  evaluate the degree of diversity of the generated responses by calculating the number of distinct unigrams and bigrams in the generated responses. We also evaluate each generated response by calculating BLEU score with all reference responses, and use the highest BLEU score to represent the quality of generated response. The average of all highest BLEU score in the testing set is named AH-BLEU. 
In addition, BLEU score is calculated by using the toolkit of NLTK\footnote{https://github.com/nltk/nltk}.\\
\indent For human evaluation, given 300 randomly sampled contexts and their responses which are generated by different models, three annotators (all CS majored students) are required to give the score of those context-response pairs, e.g. 3, 2, 1 means relevant, common and no-relevant, respectively, based on the coherence of the generated responses with respect to the contexts. The mean score is the average of all scores given by the three annotators with context-response pairs generated by a model. Meanwhile, in order to get the relative score of different models, we also evaluate the ground-truth context-response pairs by human evaluation.

\begin{table*}[!t]
    \centering
    \begin{tabular}{lcccccc}
        \toprule
        \multicolumn{7}{c}{STC Datasets }  \\
        \toprule
         Model & PPL & BLEU-2(\%) & BLEU-4(\%) &DIS-1(\%) &DIS-2(\%)&AH-BLEU-2(\%)\\
         \toprule
          Transformer &28.86 &3.74 &1.37 &0.23 &0.90 &14.43\\
          RS-Word Oracle &28.91& 5.12& 2.05& 0.33& 1.25& 15.21\\
          RS-Sentence Oracle & \textbf{26.75}&\textbf{5.50} & 2.12& 0.35& 1.38& 15.52\\
          AdapBridge &29.36& 5.35& \textbf{2.17}& \textbf{0.43}& \textbf{1.74}&\textbf{16.38} \\
          \toprule
         \multicolumn{7}{c}{Reddit Datasets }  \\
         \toprule
         Model & PPL & BLEU-2(\%) & BLEU-4(\%) &DIS-1(\%) &DIS-2(\%)&AH-BLEU-2(\%)\\
         \toprule			
          Transformer & 40.83& \textbf{3.99}& 0.77& 0.79& 2.91& 7.03\\
          RS-Word Oracle &43.11& 3.78&0.81 &1.42 &5.19 & 7.43\\
          RS-Sentence Oracle &\textbf{40.72}& 3.49& 0.76& 1.33&5.08 &7.05\\
          AdapBridge &48.01 &3.56 & \textbf{0.83}& \textbf{1.56}& \textbf{5.56}&\textbf{7.60}\\
         \toprule
    \end{tabular}
    \caption{The metric based evaluation results on STC and Reddit datasets. DIS represent the distinct score and AH-BLEU-2 represent the average of all highest BLEU-2 score.}
    \label{tab:metricresult}
\end{table*}

\subsection{Experimental Results}
In this section, we demonstrate our experimental results on the two public datasets.
\subsubsection{Metric-based Evaluation}
Table ~\ref{tab:metricresult} shows the quantitative evaluation results. From this table, we can see that models with switch mechanism, such as RS-Word, RS-Sentence and AdapBridge, outperform the traditional Transformer-based model in terms of BLEU, Distinct-1 and Distinct-2 evaluations. The results show that the switch mechanism plays an important role in the dialogue generation task.

RS-Word and RS-Sentence both replace the ground truth tokens  by the generated tokens with a random scheduled sampling. However, their performances are both worse than our proposed model, as our model considers the relevance between the generated words and the ground truth with word-level matching scores. For the BLEU score on STC dataset as an example, the BLEU-4 score of AdapBridge is 2.17, which is better than that of RS-Word and RS-Sentence, i.e., 2.05 and 2.12. Especially, our model achieves best AH-BLEU-2 score on both two datasets, which is the significant performance gains. It shows that the responses of our model have higher quantity than other baselines.

The diversity of responses can be evaluated by Distinct score. As shows in Table ~\ref{tab:metricresult}, our AdapBridge achieves significant performance gains. Take the results of Reddit in Table ~\ref{tab:metricresult} as an example, the proposed AdapBridge model improves the Transformer, RS-Word and RS-Sentence models by 2.65, 0.37 and 0.48 Distinct-2 points, respectively. We can also note that our model has the highest Distinct score on both STC and Reddit datasets, which indicates that our model can generate more diverse response and avoid generating common responses. 
In summary, our proposed AdapBridge model has the ability to generate high quality and diverse responses, compared with baselines. We also conducted the significant test, and the result shows that the improvements of our model are significant on both two datatests, i.e., $p-value<0.01$.

\begin{table}[!t]
    \centering
    \begin{tabular}{lcccc}
        \toprule
        \multicolumn{5}{c}{STC Datasets }  \\
        \toprule
         Model & 3(\%) & 2(\%) & 1(\%) &Mean \\
         \toprule
          Ground Truth &82.23& 13.56& 4.23 & 2.78\\
          \hline
          Transformer  &48.67 &23.11 & 28.22&2.20\\
          RS-Word   & 56.33& 20.89& 22.78 &2.34\\
          RS-Sentence  & 55.33& 22.67&22.00 &2.33\\
          AdapBridge  &59.56& 24.33&16.11 & \textbf{2.43} \\
          \toprule
         \multicolumn{5}{c}{Reddit Datasets }  \\
         \toprule
         Model & 3(\%) & 2(\%) & 1(\%) &Mean \\
         \toprule
          Ground Truth & 79.00& 15.67&5.33 &2.74\\
          \hline
          Transformer  & 49.78& 21.89& 28.33&2.21 \\
          RS-Word  & 52.44& 23.00& 24.56&2.28\\
        RS-Sentence &53.11&24.67&22.22&2.31\\
          AdapBridge  &55.67 & 28.33& 16.00&\textbf{2.40} \\
         \toprule
    \end{tabular}
    \caption{The human evaluation on STC and Reddit.}
    \label{tab:huameval}
\end{table}
\subsubsection{Human Evaluation}
The human evaluation results are shown in Table ~\ref{tab:huameval}. The percentages of relevant, common and no-relevant are given to evaluate the quality of responses generated by different models. From the results we can see that our AdapBridge gets the highest score in human evaluation. Take the STC as an example, compared with Transformer, RS-Word Oracle and RS-Sentence Oracle, the AdapBridge achieves performance gains 22.38\%, 5.73\%, 7.65\% on the relevant score. For the Mean score, we can observe that our AdapBridge generates  the most relevant responses, while generates less no-relevant responses, which indicates that the responses generated by our model are attractive to annotators. We also conducted the significant test, and the result shows that the improvements of our model are significant on both two datatests, i.e., $p-value<0.01$.

\subsection{Case study}

In this section, we conduct case studies to show our model can generated more relevant and diverse responses than other baseline models. 

We give two examples as in Table ~\ref{tab:casestudy}. In the Example 1, the response of Transformer is ``Is this a fish or a fish?", which is an unreasonable sentence, as the words in the positions of two ``fish" should be different according to common sense. For the response of RS-Word, we can see that it repeats ``want to eat fish" twice, which is a part of the context. Although the response of RS-Sentence ``How lovely! I want to eat, too." is relevant, it conforms to a common response paradigm, such as ``how $\cdots$, I want it, too" or ``what's this, I want to $\cdots$". If the context contains food, animal, locations etc., such  responses all seem appropriate, which is not attractive 
to human. While the response generated by our AdapBridge is more specific and relevant, i.e. ``What kind of fish is this?". We can also see the similar phenomenon in the Example 2 of Table ~\ref{tab:casestudy}, with the context ``Venice, the city of water, the city of dreams.", Transformer repeats the same content that comes from the context, and the responses of  RS-Word and RS-Sentence are both common responses as mentioned above. Compared with responses generated by baseline models, response of AdapBridge ``I want to go here with my parents." is  more relevant and attractive. Therefore, those results indicate that our proposed model can generate high quality and attractive responses with the adaptive switch mechanism.
\begin{CJK*}{UTF8}{gbsn}
\begin{table*}
\centering
\scalebox{0.9}{
\newcommand{\tabincell}[2]{\begin{tabular}{@{}#1@{}}#2\end{tabular}}
\begin{tabular}{lll} 
\toprule
\multicolumn{3}{c}{Example 1}\\
\toprule
context &  \tabincell{l}{你不是天天嚷着要吃鱼嘛?鱼来了} &  \tabincell{l}{Don't you want to eat fish every day? The fish is coming.} \\ 
Transformer &  \tabincell{l}{这是鱼还是鱼啊?} & \tabincell{l}{Is this a fish or a fish?}\\ 
RS-Word &  \tabincell{l}{要吃鱼，要吃鱼。} & \tabincell{l}{Want to eat fish, want to eat fish}\\ 
RS-Sentence &  \tabincell{l}{好可爱！我也想吃。} & \tabincell{l}{ How lovely! I want to eat, too.}\\ 
AdapBridge &  \tabincell{l}{这是什么鱼?} & \tabincell{l}{ What kind of fish is this?}\\ 
\toprule
\multicolumn{3}{c}{Example 2}\\
\toprule
context &  \tabincell{l}{水城威尼斯，梦里的城市。} &  \tabincell{l}{Venice, the city of dreams.}\\ 
Transformer &  \tabincell{l}{梦里的城市，梦里的城市。} & \tabincell{l}{ City in dreams, city in dreams.}\\ 
RS-Word &  \tabincell{l}{ 好美的城市，好想去。} & \tabincell{l}{Beautiful city. I really want to go.}\\ 
RS-Sentence &  \tabincell{l}{这是哪呀！我也想去。} & \tabincell{l}{Where is this? I want to go too.}\\ 
AdapBridge &  \tabincell{l}{想和我的爸妈一起去。} & \tabincell{l}{I want to go here with my parents.}\\ 
\bottomrule
\end{tabular}
}
\caption{Two examples of generated responses on STC.} 
\label{tab:casestudy}
\end{table*}
\end{CJK*}

\subsection{AdapBridge on NMT}
The method we propose can also be adapted for the neural machine translation(NMT) in an easy way. With this task we want to investigate if AdapBridge could be helping to improve the performance of NMT which is a classic natural language generation task. We perform experiments on the WMT'14  English$\to$German(En$\to$De) datasets, which contains 3,900,502, 39,414, 3,003 sentences for training, validation and testing sets, respectively. We train the Transformer-based model with the same setting described in Section ~\ref{ch:setting}, and then we measure the translation quality with BLEU. The evaluation results are listed in Table ~\ref{tab:translation}.

From the results, we can see that our method can also achieve significant performance gains, and improve the Transformer-based model by 0.95 BLEU-4 points on average. For the BLEU-2 score, our model is slightly lower than RS-Sentences model same as the results in Table ~\ref{tab:metricresult}, it can attributed to the sentence-level information of RS-Sentence Oracles. In order to analyze the gap of ground-truth and the generated sentences, we calculate the cosine similarity between the hidden representations of ground truth sentences and generated sentences, with a trained Bert model ~\cite{wolfetal2020transformers}, and get the similarity score 0.96 and 0.81 on WMT'14 and Reddit datastes, respectively. At the same time, we can also notice that BLEU score of NMT is much higher than that of dialogue generation task. The results of overlap measures indicate the severity of the exposure bias problem in dialogue generation as analyzed in Section ~\ref{ch:intro}.
\begin{table}[!t]
    \centering
    \begin{tabular}{lcc}
    \toprule
         Model  & BLEU-2(\%) & BLEU-4(\%)\\
        \toprule
         Transformer&43.40&26.43  \\
         RS-Word&43.66 &26.84  \\
         RS-Sentence&\textbf{44.08}&27.21\\
         AdapBridge&43.99&\textbf{27.38} \\
    \toprule
    \end{tabular}
    \caption{BLEU scores on En$\to$De NMT task.}
    \label{tab:translation}
\end{table}

\section{Conclusion}
In this paper, we propose a novel adaptive switch mechanism with word-level matching scores to solve the problem of exposure bias for the dialogue generation task, named AdapBridge. Our core idea is to utilize the word-level matching scores to determine the input is from ground truth or from prediction at each step of training. Experimental results show that our model significantly outperforms previous baseline models. Further analysis on NMT also indicates that our model can achieve significant improvement on different generation tasks. In future work, we plan to further design different scoring methods, i.e. Bert score or BLEU, to guide the model selects better words. It is also interesting to extend our AdapBridge model to other generation tasks, such as abstractive summarization.

\section*{Acknowledgements}
This work is supported by the Beijing Academy of Artificial Intelligence (BAAI), and the National Natural Science Foundation of China (NSFC) (No.61773362).

\bibliography{anthology,custom}
\bibliographystyle{acl_natbib}




\end{document}